%% file: main.tex
\documentclass[10pt,twocolumn,letterpaper]{article}

\usepackage{cvpr}      
\usepackage{algorithm}
\usepackage{algorithmic}
\usepackage{times}
\usepackage{epsfig}
\usepackage{amssymb,enumitem}
\usepackage{graphicx}
\usepackage{amsmath}
\usepackage{bbding}
\usepackage{amsthm}
\usepackage{booktabs}
\usepackage{array,multirow}
\usepackage{color}
\usepackage{xcolor}
\usepackage{makecell}
\usepackage{epsfig}
\usepackage{enumerate}

\definecolor{cvprblue}{rgb}{0.21,0.49,0.74}
\usepackage[pagebackref,breaklinks,colorlinks,allcolors=cvprblue]{hyperref}

\def\paperID{9826} 
\def\confName{CVPR}
\def\confYear{2025}

\title{High-fidelity 3D Object Generation from Single Image with RGBN-Volume Gaussian Reconstruction Model}

\author{Yiyang Shen \and Kun Zhou \and He Wang \and Yin Yang \and Tianjia Shao
}
\author{Yiyang Shen$^1$ \quad Kun Zhou$^1$ \quad He Wang$^2$\quad Yin Yang$^3$ \quad Tianjia Shao$^{1\dag} $\\
$^1$State Key Lab of CAD\&CG, Zhejiang University \\$^2$AI Centre, University College London  \quad $^3$University of Utah
}

\begin{document}

\twocolumn[{%
\renewcommand\twocolumn[1][]{#1}%
\maketitle

\thispagestyle{empty}
\begin{center}
     \includegraphics[width= 1.0\linewidth]{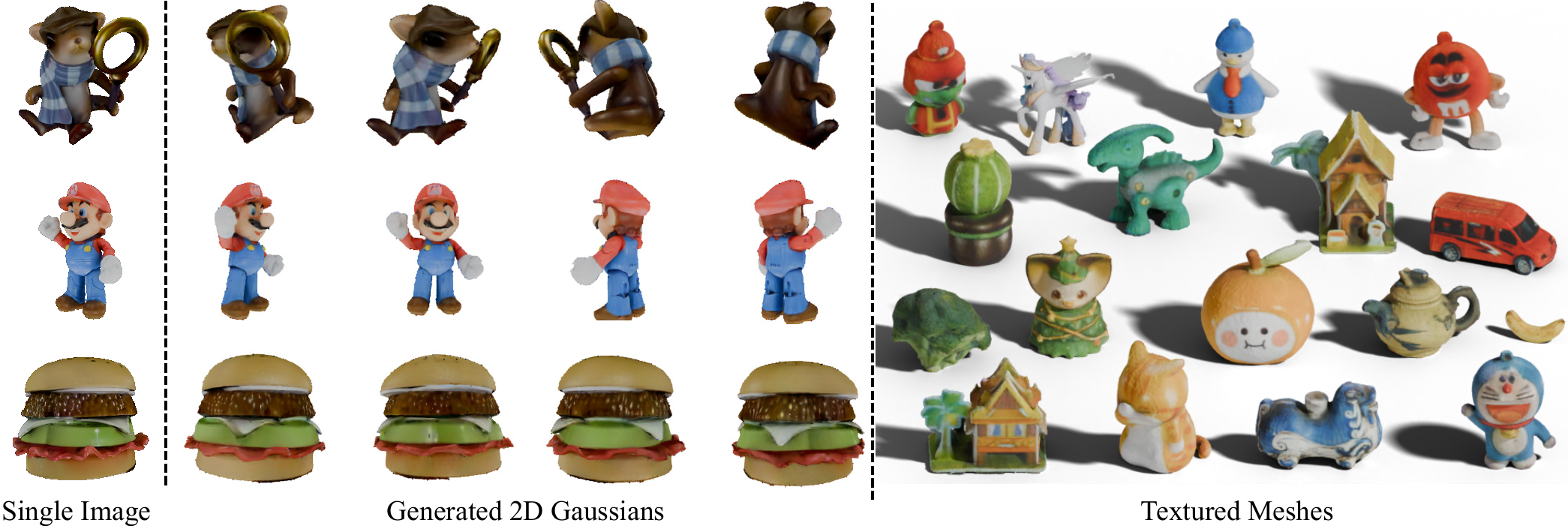}
     \captionof{figure}{GS-RGBN is an RGBN-volume Gaussian reconstruction model that generates high-quality 2D Gaussians (middle) using a single image (left). The textured meshes can be reconstructed from the generated 2D Gaussians optionally (right).
        }
    \label{fig:3D_scence}
\end{center}
}]

\let\thefootnote\relax\footnotetext{$^{\dag}$ Corresponding author.}

\begin{abstract}
%

Recently single-view 3D generation via Gaussian splatting has emerged and developed quickly. They learn 3D Gaussians from 2D RGB images generated from pre-trained multi-view diffusion (MVD) models, and have shown a promising avenue for 3D generation through a single image. Despite the current progress, these methods still suffer from the inconsistency jointly caused by the geometric ambiguity in the 2D images, and the lack of structure of 3D Gaussians, leading to distorted and blurry 3D object generation. In this paper, we propose to fix these issues by \textbf{GS-RGBN}, a new RGBN-volume Gaussian Reconstruction Model designed to generate high-fidelity 3D objects from single-view images. Our key insight is a structured 3D representation can simultaneously mitigate the afore-mentioned two issues. To this end, we propose a novel hybrid Voxel-Gaussian representation, where a 3D voxel representation contains explicit 3D geometric information, eliminating the geometric ambiguity from 2D images. It also structures Gaussians during learning so that the optimization tends to find better local optima. Our 3D voxel representation is obtained by a fusion module that aligns RGB features and surface normal features, both of which can be estimated from 2D images. Extensive experiments demonstrate the superiority of our methods over prior works in terms of high-quality reconstruction results, robust generalization, and good efficiency.

\end{abstract}


\section{Introduction}
Crafting 3D assets from 2D images has broad applications in fields such as virtual reality (VR), augmented reality (AR), industrial design, gaming, and animation.
Recently, significant attention has been focused on utilizing only a single image to generate a 3D object with superior shapes and textures as a subtopic.
However, the persisting challenge arises due to the inherent geometric ambiguity and limited information provided in single-view images.
%
%

Emerging multi-view diffusion (MVD) models \cite{rombach2022high,saharia2022photorealistic} present a potential solution to address the above information scarcity. 
These models extend one image to multi-view images, thus providing more comprehensive information from different viewpoints for 3D object generation.
The pioneering work (Dreamfusion) \cite{poole2022dreamfusion} and following works \cite{chen2023fantasia3d,deng2023nerdi,melas2023realfusion,ouyang2023chasing,qian2023magic123,sun2023dreamcraft3d,tang2023dreamgaussian} propose score distillation sampling (SDS) and some variants, which directly leverage multi-view images (or their 3D prior knowledge) generated by pre-trained MVD models to optimize a 3D parametric model (e.g., NeRF \cite{deng2023nerdi}, SDF \cite{cheng2023sdfusion}, point clouds \cite{nichol2022point} and 3D Gaussian Splatting \cite{tang2023dreamgaussian,chen2024text}).
However, these MVD images exhibit significant inconsistency across different viewpoints and generate view-inconsistent 3D objects.

To mitigate this issue, another group of works \cite{liu2023zero,shi2023mvdream,wang2023imagedream,long2024wonder3d,voleti2025sv3d} resort to leveraging additional information, \eg camera embeddings \cite{liu2023zero}, text embeddings \cite{shi2023mvdream} and epipolar constraints \cite{huang2024epidiff}, to fine-tune the pre-trained MVD models. Despite these improvements, the fine-tuned MVD images still fail to meet the demand for directly reconstructing 3D models with consistent details. Additionally, the per-shape optimization process requires thousands of iterations for each object, leading to slow 3D reconstruction. It raises a question - instead of primarily focusing on fine-tuning MVD models to enhance image consistency for the per-shape optimization process, can we develop an end-to-end neural network that directly learns from inconsistent MVD images to generate view-consistent 3D objects without relying on intricate optimization iterations?

To answer this question, recent methods \cite{zou2024triplane,tang2025lgm,xu2024grm,liu2024one,liu2024one++}, pioneered
by the large reconstruction model (LRM) \cite{hong2023lrm}, employ diverse neural networks (e.g., transformer \cite{xu2024grm} and U-Net \cite{tang2025lgm}) that directly learn from inconsistent MVD images to generate 3D models.
The generated 3D models are subsequently used to render per-view images, which supervise the training process via a rendering loss between the rendered images and ground-truth.
Especially, 3D Gaussian Splatting (3DGS) \cite{kerbl20233d} has emerged as the predominant 3D representation in most feed-forward models \cite{tang2025lgm,xu2024grm,zhang2025gs,shen2024gamba}, owing to its exceptional quality of novel view synthesis and fast rendering speed, replacing previous 3D representations like NeRF \cite{mildenhall2021nerf}.
However, the direct learning of 3D Gaussians from 2D images for high-fidelity 3D object generation remains a challenge due to the spatially unstructured nature of 3DGS \cite{zou2024triplane,zhang2024gaussiancube} and the inherent geometric ambiguity in input 2D RGB images, leading to distorted and blurry 3D object generation.

To this end, we propose GS-RGBN, an RGBN-volume Gaussian reconstruction model capable of fast and high-quality rendering and reconstruction for 3D objects within a few seconds (see Fig. \ref{fig:3D_scence}). GS-RGBN implements two key insights: 
first, unlike traditional methods that employ 2D convolutions to encode image features and decode corresponding per-pixel 3D Gaussian attributes in 2D planes, 
we propose a novel hybrid Voxel-Gaussian model where each Gaussian is constrained within a voxel grid, where each voxel contains the projected 2D image features.
It achieves a spatial correspondence between the 3D location of each Gaussian and its corresponding 2D projected image features, permitting the use of standard 3D convolutions to effectively capture correlations among neighboring Gaussians for generalizable 3D representation learning.
Second, normals offer crucial geometric cues for recovering intricate details that are lost due to the inherent geometric ambiguity in previous RGB-only 3D reconstruction methods.
%
Therefore, we propose a simple but effective cross-volume fusion (CVF) module with multiple cross-attentions to leverage the complementary semantic and geometric information from RGB and normal images for feature-level fusion. As a result, the fused features can be utilized to enhance the geometric intricacies of reconstructed objects.
Moreover, we adopt 2D Gaussian \cite{huang20242d} as 3D representation, instead of the widely used 3D Gaussian \cite{kerbl20233d}, thus ensuring consistent geometric representation and intrinsic modeling of surfaces.
In summary, our contributions are as follows:
\begin{itemize}
\item We propose a novel RGBN-volume Gaussian reconstruction model, called GS-RGBN, to generate high-quality 3D assets from single-view images in just a few seconds.
\item We propose a hybrid Voxel-Gaussian model that provides a well-structured 3D grid representation for generalizable 3D learning of unstructured Gaussians.
\item We propose a simple but effective cross-volume fusion (CVF) module for feature-level RGB and normal fusion to recover high-fidelity geometry.
\item Extensive experiments demonstrate that our method outperforms existing paradigms in both geometry reconstruction and novel view synthesis.
\end{itemize}

\begin{figure*}[!t] \centering
	\includegraphics[width=1\linewidth]{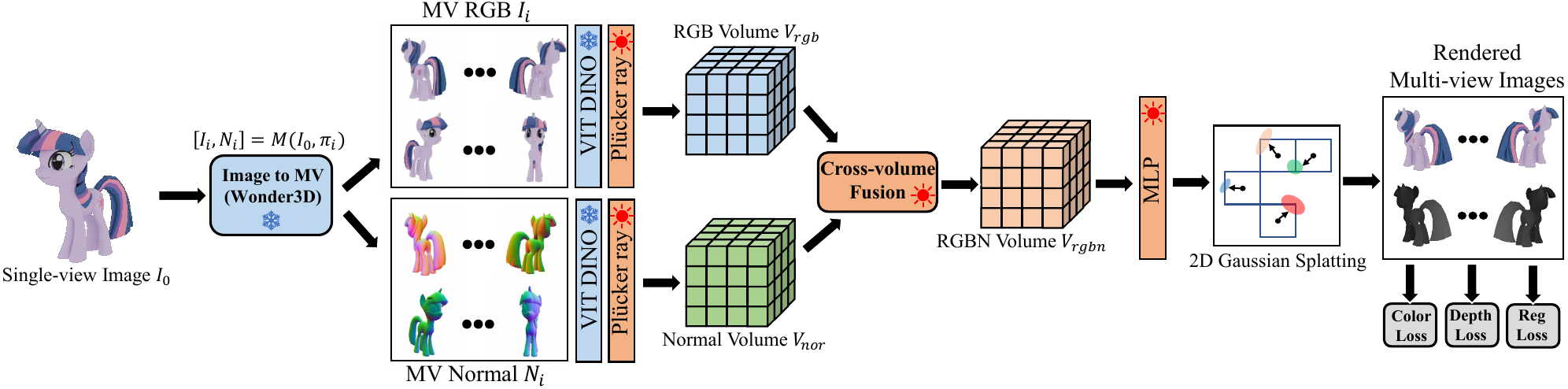}
	\caption{
 The overview of our paradigm. 
 Given a single image of a 3D object, we first input it into an off-the-shelf multi-view diffusion model (Wonder3D \cite{long2024wonder3d}) to obtain two sets of multi-view normal and RGB images, which are used to build the hybrid Voxel-Gaussian model.
 Especially, we input these images to pre-trained VIT DINO models \cite{caron2021emerging} and lift extracted 2D DINO features to build two 3D feature volumes, i.e., RGB feature volume $V_{rgb}$ and normal feature volume $V_{nor}$ modulated by Plücker rays (Sec. \ref{SecM.1}). 
 Next, a feature-level cross-volume fusion (CVF) module is capable of effectively fusing the RGB and normal volumetric features to obtain the fine-grained fused RGBN feature volume $V_{rgbn}$ (Sec. \ref{SecM.2}). 
 Finally, we use several MLPs for decoding $V_{rgbn}$ to regress 2D Gaussian primitives for novel view rendering (Sec. \ref{SecM.3}). 
 Notably, the training process is supervised by color, depth and regularization loss functions (Sec. \ref{SecM.4}). 
    }
	\label{fig:network framework}
\end{figure*}
\section{Related Work}
Creating 3D assets from only single-view images is an ill-posed problem that has received persistent attention.
Inspired by the successes of growing diffusion models for multi-view image generation \cite{rombach2022high,zhang2023adding,liu2023zero}, current methods leverage multi-view images (or their 3D priors) from pre-trained MVD models to reconstruct 3D objects. 
Based on their distinct utilization of MVD models, these methods can be divided into three categories: optimization-based, fine-tune-based, and feed-forward methods.

\textbf{Optimization-based 3D generation.}
Starting with Dreamfields \cite{jain2022zero} and Dreamfusion \cite{poole2022dreamfusion}, optimization-based approaches \cite{chen2023fantasia3d,deng2023nerdi,melas2023realfusion,ouyang2023chasing,qian2023magic123,sun2023dreamcraft3d,tang2023dreamgaussian,melas2023realfusion,zhou2024gala3d,lin2023magic3d,metzer2023latent,chen2023fantasia3d,haque2023instruct,ma2023geodream,zhu2023hifa,bahmani20244d,lorraine2023att3d} employ score distillation sampling (SDS) or some variants for the pre-trained MVD models to optimize a 3D parametric model, such as NeRF \cite{deng2023nerdi,metzer2023latent}, SDF \cite{cheng2023sdfusion}, point clouds \cite{nichol2022point,melas2023pc2} and 3D Gaussian Splatting \cite{tang2023dreamgaussian,chen2024text}. 
For example, DreamGaussian \cite{tang2023dreamgaussian} first adopts SDS-based 2D diffusion priors to optimize 3D Gaussians, which are refined by the following UV-space texture refinement stage. 
Gaussiandreamer \cite{yi2024gaussiandreamer} bridges the abilities of 3D and 2D diffusion models via the Gaussian splatting representation for fast text-to-3D.
These methods avoid the dilemma of using 3D data for training, yet the lack of consistency in different viewpoints among MVD images leads to suboptimal generation of 3D objects.

\textbf{Fine-tune-based 3D generation.} 
Inspired by the successes of fine-tuned approaches \cite{hu2021lora,zhang2023adding,rombach2022high}, fine-tune-based approaches \cite{liu2023zero,shi2023mvdream,wang2023imagedream,long2024wonder3d,voleti2025sv3d,chen2024text,liang2024luciddreamer,ling2024align,wang2024prolificdreamer,cheng2023progressive3d} first add conditional controls, e.g., camera embeddings \cite{liu2023zero,shi2023zero123++}, text embeddings \cite{shi2023mvdream} and epipolar constraints \cite{huang2024epidiff}, to fine-tune pre-trained MVD models for ensuring consistency across multi-view images.
Similar to optimization-based ones, they use fine-tuned MVD images to optimize a 3D parametric model.
For example, the pioneering work Zero-1-to-3 \cite{liu2023zero} learns controls of the relative camera viewpoint to provide fine-tuned MVD models with the ability to perceive diverse views.
Follow-up works Mvdream \cite{xu2023dmv3d} and Imagedream \cite{wang2023imagedream} add encoded text/image features as controls to fine-tune the diffusion model, enhancing texture details of generated 3D objects.
Despite significant investments in time and resources to fine-tune MVD models, the fine-tuned MVD images still exhibit inconsistency, leading to blurry and distorted 3D object generation.

\textbf{Feed-forward 3D generation.}
Inspired by the successes of the Large Reconstruction Model (LRM) \cite{hong2023lrm}, recent feed-forward single-view 3D generation approaches are proposed.
Considering the challenges of fine-tuning MVD models to improve view consistency, feed-forward approaches \cite{zou2024triplane,tang2025lgm,xu2024grm,liu2024one,liu2024meshformer,shen2024gamba,melas20243d,zhang2025gs,xu2024agg} directly learn such inconsistent MVD images to optimize 3D models. 
%
Especially, recent feed-forward methods \cite{zou2024triplane,tang2025lgm,xu2024grm,zhang2025gs} commonly employ 3D Gaussian Splatting as the preferred 3D representation due to its rapid rendering speed and superior rendering quality, compared with previous 3D representations (like NeRF \cite{mildenhall2021nerf}).
Our GS-RGBN is also a feed-forward 3D reconstruction paradigm. 
It deviates from traditional feed-forward models by a 3D-native structure, i.e., a hybrid Voxel-Gaussian model, to achieve generalizable
3D learning of unstructured Gaussians, and a cross-volume fusion module to effectively fuse RGB and normal features for enhancing the geometry of reconstructed 3D objects.

\section{Method}\label{proposedmethod}
As shown in Fig. \ref{fig:network framework}, GS-RGBN takes as input a single image of a 3D object into the MVD model Wonder3D \cite{long2024wonder3d} to obtain two sets of multi-view RGB and normal images, which are used to generate voxel-based 2D Gaussians for high-fidelity 3d object generation.
In the following sections, we first introduce how to build a hybrid Voxel-Gaussian model using multi-view RGB and normal images (Sec. \ref{SecM.1}). 
Then, we propose a simple but effective feature-level cross-volume fusion module that fuses the RGB and normal volumes to reproduce a fine-grained RGBN volume, aligning both crucial semantic (RGB) and geometric (normal) cues for subsequent 2D Gaussian decoding (Sec. \ref{SecM.2}). 
Next, we describe how to decode the RGBN volume to generate high-quality 2D Gaussians for novel view rendering and high-quality shape reconstruction (Sec. \ref{SecM.3}). 
Lastly, we will present the training objective, which includes the supervision of color, depth and regularization loss functions (Sec. \ref{SecM.4}).

\subsection{Hybrid Voxel-Gaussian}\label{SecM.1}
3D Gaussian splatting \cite{kerbl20233d} offers good rendering speed and quality compared with previous 3D representations (e.g., mesh \cite{wang2018pixel2mesh}, point clouds \cite{fan2017point}, and NeRF \cite{mildenhall2021nerf}). 
However, if the 2D views are highly inconsistent, it can lead to unstable and cumbersome training and generating objects with subpar geometry and blurry textures~\cite{tang2025lgm,xu2024grm,zhang2025gs,zou2024triplane,tang2023dreamgaussian} (Fig. \ref{fig:RGB_C} and \ref{fig:GEO_C}).
%
Therefore, we propose a hybrid Voxel-Gaussian model that builds a structured 3D voxel grid, where each voxel contains projected 2D image features for decoding per-voxel Gaussians. It establishes correspondences between the 3D positions of each Gaussian and the corresponding projected 2D image features, which further enables 3D convolutions to effectively capture the correlations among neighboring Gaussians, leading to a generalizable 3D representation.

Given a single image $I_{0}$, we first feed it into a multi-view diffusion model (Wonder3D \cite{long2024wonder3d}) $M$ which generates multi-view RGB and normal images $[I_{i},N_{i}] = M(I_{0},\pi_{i})$ of a target 3D object with diverse camera poses $P=[\pi_{i}]$. 
Then, we use these multi-view RGB/normal images with corresponding camera poses to build RGB/normal volumes. 
Concretely, we feed RGB images into a pre-trained robust VIT DINO model \cite{caron2021emerging} to obtain corresponding per-view image feature maps. 
Following \cite{tang2025lgm,xu2023dmv3d,xu2024grm,chen2024lara}, we then use the Plücker ray embedding \cite{sitzmann2021light} to encode corresponding camera poses. 
Especially, Plücker ray embedding provides a distinctive representation of rays in 3D space, formed by computing the cross-product of the camera's position vector (rays' origin) $o_{i}$ and the ray's directional vector $d_{i}$. 
Subsequently, we inject such Plücker ray embedding into the per-view image feature maps via the adaptive layer norm \cite{peebles2023scalable} to obtain the fused feature map that contains information on per-view images and corresponding viewpoints, which can be formulated as
\begin{equation}
\label{RGBP}
f_{i} = \operatorname{Norm}(c_{i},o_{i}\times d_{i},d_{i})
\end{equation}
where $f_{i}$ and $c_{i}$ denote the fused feature and RGB feature for pixel $i$, respectively.
The fused features are back-projected along each ray into per-view 3D feature volumes ${V}_{i},i=1,2,...,n$, and the final RGB feature volume $V_{rgb}\in R^{W \times W \times W \times C}$ is obtained by averaging the features at the same position across per-view volumes $V_{rgb}=avg(V_{1},V_{2}...,V_{n})$.
Notably, building normal volume $V_{nor}$ is the same as the above RGB volume building process.


\begin{figure}[!t] \centering
	\includegraphics[width=1\linewidth]{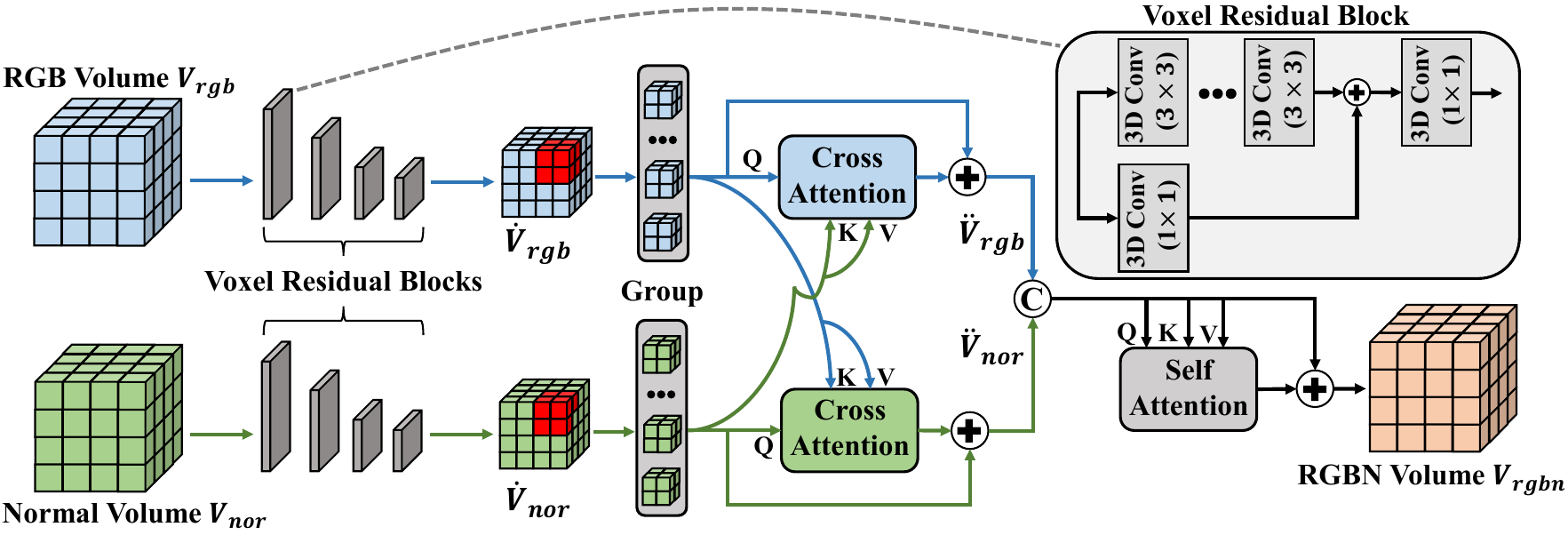}
	\caption{The illustration of the structure of the cross-volume fusion (CVF) module. 
    }
	\label{fig:CVF}
\end{figure}

\subsection{Cross-volume Fusion}\label{SecM.2}
Traditional feed-forward works \cite{tang2025lgm,xu2024grm,zou2024triplane,liu2024one,liu2024one++} only extract 2D RGB feature maps to reconstruct 3D objects.
However, unlike normal images that explicitly encode geometric information, the reconstruction of 3D objects from RGB images only captures semantic details and suffers from insufficient geometric details (see Fig. \ref{fig:GEO_C}).
It is reasonable to leverage both RGB and normal images that offer complementary semantic and geometric information for high-quality 3D object generation.
Thus, we propose a simple but effective feature-level cross-volume fusion (CVF) module to fuse RGB and normal volumetric features to build a fine-grained RGBN volume.
%

We now present our CVF module (see Fig. \ref{fig:CVF}) that contains four voxel residual blocks, two cross-attention blocks, and one self-attention block. 
We first use four voxel residual blocks (VRBs) with feature channels $[512,256,128,32]$, extended from the 2D residual blocks \cite{he2016deep}, 
%
to downsample RGB and normal volumetric features, leading to reduced memory overhead.
Concretely, these volumetric features are fed into the main path that contains a set of 3$\times$3 3D convolutional layers followed by LeakyReLU, and the shortcut path that contains a 1$\times$1 3D convolutional layer followed by LeakyReLU.
The additional shortcut path can effectively solve the gradient vanish problem, thus leading to stable training for deep learning models.
Then, extracted features of such two branches are added, integrating lost value information from shadower blocks to deeper ones, and passed through a 1$\times$1 3D convolutional layer followed by a normal layer to obtain the downsampled feature volumes ($\dot{V}_{rgb}$ and $\dot{V}_{nor}$).

Regarding the complementary nature of semantic and geometric features, two cross-attention blocks ($\operatorname{CA}_{s}$ and $\operatorname{CA}_{g}$) are proposed to dynamically capture the correlations between RGB and normal volumetric features.
Before it, we unfold 3D feature volumes ($\dot{V}_{rgb}$ and $\dot{V}_{nor}$) with a resolution of $32\times32\times32$ into $G=16$ groups along each axis \cite{chen2024lara}, which effectively reduces memory and time overheads while preserving model performance.
Specially, we first map the groups of $\{\dot{V}_{rgb}^{g}\}^{G}_{g=1}$ into a query $Q$ and the groups of $\{\dot{V}_{nor}^{g}\}^{G}_{g=1}$ into keys $K$ and values $V$ and pass though the RGB-guided cross-attention block ($\operatorname{CA}_{s}$) to obtain RGB-guided fused volume $\ddot{V}_{rgb}$. 
Notably, another normal-guided cross-attention block $\operatorname{CA}_{g}$ with the same network structure as $\operatorname{CA}_{s}$ is adopted to map $\dot{V}_{nor}$ into a query as guidance to obtain normal-guided fused volume $\ddot{V}_{nor}$, making the fusion more focused on geometric (normal) information.
%

Finally, we concatenate $\ddot{V}_{rgb}$ and $\ddot{V}_{nor}$ and input the results into a self-attention block $\operatorname{SA}$ to effectively balance the weights assigned to semantic and geometric information, aggregating them to obtain the ultimate RGBN volume denoted as $V_{rgbn}$.
%
%
The whole fusion process can be formulated as
\begin{equation}
\label{c1}
\ddot{V}_{rgb}^{g} = \operatorname{CA}_{s}(\operatorname{LN}(Q=\{\dot{V}^{g}_{rgb}\}),K,V=\{\dot{V}^{g}_{nor}\})+\dot{V}^{g}_{rgb}
\end{equation}
\begin{equation}
\label{c2}
\ddot{V}_{nor}^{g} = \operatorname{CA}_{g}(\operatorname{LN}(Q=\{\dot{V}^{g}_{nor}\}),K,V=\{\dot{V}^{g}_{rgb}\})+\dot{V}^{g}_{nor}
\end{equation}
\begin{equation}
\label{c3}
\ddot{V}^{g}_{rgbn} = \ddot{V}^{g}_{rgb} \oplus \ddot{V}^{g}_{nor}
\end{equation}
\begin{equation}
\label{c4}
V_{rgbn}^{g} = \operatorname{SA}(Q,K,V=\{\ddot{V}^{g}_{rgbn}\}) + \ddot{V}^{g}_{rgbn}
\end{equation}
where $CA_{(.)}$, $SA$, $LN$, and $\oplus$ represent cross-attention blocks, self-attention blocks, layer norms, and concatenation, respectively. And $g$ denotes the index of the group.

\subsection{2D Gaussian Generation}\label{SecM.3}
Unlike widely used 3D Gaussians, 2D Gaussians have been proven to ensure consistent representation of geometry and intrinsic modeling of surfaces \cite{huang20242d}.
Thus, we adopt 2D Gaussian Splatting to effectively reconstruct geometry surfaces from inconsistent multi-view images.
Concretely, each 2D Gaussian is defined by a center $x \in R^{3}$, a scaling factor $s \in R^{2}$ and a rotation factor $q \in R^{4}$ to control the shape of the 2D Gaussian.
Additionally, an opacity value $\alpha \in R$ and a spherical harmonics (SH) coefficient $sh \in R^{C}$ are maintained to incorporate view-dependent effects in the rendering process.
For the RGBN volume $V_{rgbn}$, we query features $V_{rgbn}^{i}$ from the $i \text {-th }$ voxel and adopt a set of MLPs $\phi_{g}$ to decode the attributes of the per-voxel 2D Gaussians:
\begin{equation}
\label{mlp}
(\Delta x_{i},s_{i},q_{i},\alpha_{i},sh_{i}) = \phi_{g}(V_{rgbn}^{i})
\end{equation}
where $\Delta x_{i} \in {[ -1,1]}^{3}$ denotes an offset vector, incorporating a sigmoid activation function. 
The final position of 2D Gaussian in voxel $v_{i}$ can be computed by $x_{i} = v_{i}+r\bullet \Delta x_{i}$, where $r$ represents the maximum movement range of the primitive. 
It enables each Gaussian to be positioned in close proximity to the corresponding local voxel center, effectively representing adjacent regions that are required by corresponding 2D projected pixels.



\textbf{Rendering.} We take advantage of Gaussian splatting \cite{kerbl20233d,huang20242d} to perform image rendering at any novel viewpoint. 
Following the original rasterization process \cite{huang20242d}, we further incorporate the $z$ value and normal information of 2D Gaussians to obtain depth and normal maps.
Notably, several methods \cite{chung2024depth,charatan2024pixelsplat} based on 3D Gaussian splatting directly utilize the $z$ value of 3D Gaussians as their depths and employ alpha blending technique to generate final depth maps.
However, these predicted depth maps suffer from inaccuracies and low quality due to the varying depths presented by 3D Gaussians when a ray passes through the entire ellipsoid. The varying depths cannot be simply treated as the $z$ value of the center.
To solve this problem, 2D Gaussians explicit ray-splat intersection, where the pixel's depth is obtained by calculating the intersection point between the view ray and the opaque ellipsoid disc \cite{huang20242d}.

\subsection{Training Objective}\label{SecM.4}
We train the full paradigm via color $\mathcal{L}_{c}$ and depth $\mathcal{L}_{d}$ loss supervision, optimizing reconstruction objectives between rendered and ground-truth RGB/depth images. 
Additionally, a regularization loss $\mathcal{L}_{\mathrm{Reg}}$, consisting of a self-supervised distortion loss and a normal consistency loss \cite{huang20242d}, is used to improve the geometry reconstruction. 
It can be formulated as:
\begin{equation}
\label{loss_total}
\mathcal{L}_{total} = \mathcal{L}_{c}+\lambda_{d}\mathcal{L}_{d}+\lambda_{reg}\mathcal{L}_{\mathrm{reg}}
\end{equation}
\begin{equation}
\label{loss_color}
\mathcal{L}_{c} = \lambda_{1}\mathcal{L}_{\mathrm{1}}(I_{rgb}, \hat{I}_{rgb})+\lambda_{2}\mathcal{L}_{\mathrm{1}}(I_{\alpha}, \hat{I}_{\alpha})+\lambda_{3}\mathcal{L}_{\mathrm{lp}}(I_{rgb}, \hat{I}_{rgb})
\end{equation}
\begin{equation}
\label{loss_depth}
\mathcal{L}_{d} = \mathcal{L}_{\mathrm{1}}(D, \hat{D})
\end{equation}
where $I_{rgb}$/$\hat{I}_{rgb}$, $I_{\alpha}$/$\hat{I}_{\alpha}$ and $D$/$\hat{D}$ denote the ground-truth/rendered RGB, alpha and depth images. $\mathcal{L}_{\mathrm{1}}$ and $\mathcal{L}_{\mathrm{lp}}$ denote the L1 loss and VGG-based LPIPS loss \cite{zhang2018unreasonable}. 

\begin{figure*}[!ht] \centering
	\includegraphics[width=1\linewidth]{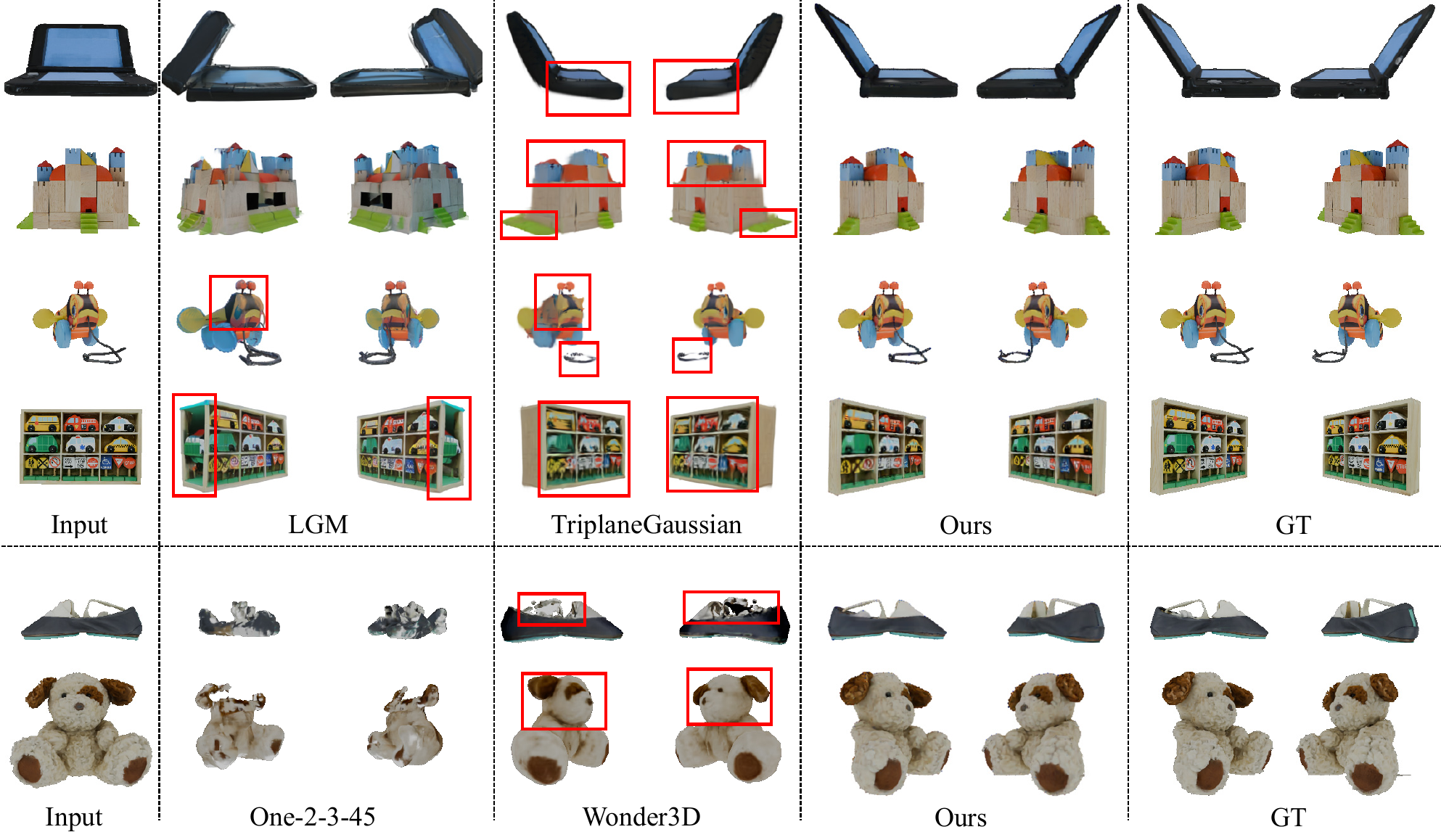}
	\caption{
Qualitative comparisons of novel view synthesis between GS-RGBN and other methods on the GSO dataset. It can be observed that the 3D objects reconstructed by our method have both high-quality and consistent details.
    }
	\label{fig:RGB_C}
\end{figure*}

\section{Experiment}
\subsection{Experimental Settings}
\textbf{Training Settings.} 
The optimization is performed using AdamW~\cite{loshchilov2017decoupled}, with an initial learning rate of $1\times10^{-5}$ and subsequently following a cosine annealing schedule with a period of 32 steps.
Our model is trained on four A100 (40G) GPUs for approximately 6.5 days, employing a batch size of four per GPU with bfloat16 precision, resulting in an effective batch size of 16.
We set $\lambda_{d}$, $\lambda_{reg}$, $\lambda_{1}$, $\lambda_{2}$, $\lambda_{3}$ to 1,0.5,1,1,0.5 in our experiments.

\textbf{Dataset.} Following \cite{zou2024triplane,liu2024one}, our model is trained on the Objaverse-LVIS dataset \cite{deitke2023objaverse} that contains 46K diverse 3D objects in 1156 categories.  
We first filter approximately 6K low-quality objects (i.e., partial scans and missing textures) and use Blender to render each remaining object to obtain the ground-truth RGB images and the depth images with a circular camera path. 
For evaluation, We adopt the most widely used Google Scanned Objects (GSO) dataset \cite{downs2022google}. 
Similar to previous methods \cite{zou2024triplane,liu2024one,tang2025lgm,xu2024grm,zheng2024mvd}, we randomly choose approximately 200 objects to render two single images (i.e., Front and side of the object) as known-view inputs per object to evaluate the performance of our method and others.

\textbf{Baselines and Metrics.} We compare GS-RGBN with recent single-view image reconstruction methods, including DreamGaussian \cite{tang2023dreamgaussian}, LGM \cite{tang2025lgm}, One-2-3-45 \cite{liu2024one}, Wonder3D \cite{long2024wonder3d} and TriplaneGaussian \cite{zou2024triplane}. 
To evaluate the single-view reconstruction quality, we adopt PSNR, SSIM, and LPIPS metrics, which quantify the similarity between rendered and ground-truth RGB/depth images from multiple views.
Besides, we adopt the Chamfer Distances (CD) to evaluate the quality of reconstructed geometries.
\begin{table}[!t]
\centering
\footnotesize
\setlength{\tabcolsep}{0.6mm}{
\begin{tabular}{c|cccc|cc}
\toprule
Method           & PSNR$\uparrow$ & SSIM$\uparrow$ & LPIPS$\downarrow$ & CD$\downarrow$&Time(g) $\downarrow$&Time(r) $\downarrow$ \\\midrule
DreamGaussian    & 17.43     &0.810     & 0.265      & 205.23     & -&28.32sec    \\
LGM              & 17.13&0.808  &0.199  &104.71   &2.45sec&0.33sec       \\
One-2-3-45       &  15.20    & 0.796     &  0.231     &95.84     &49.38sec&21.36sec     \\
Wonder3D         & 16.35     &  0.802    &  0.220     &106.37  &4.31sec&6.05 min         \\
TriplaneGaussian &16.73 & 0.793 & 0.259     &58.74 &-      &0.11sec  \\
Ours &\textbf{23.02} &\textbf{0.873} &\textbf{0.135}   &\textbf{27.49}       &4.31sec  & 0.20sec\\
\bottomrule
\end{tabular}
}
\caption{Quantitative comparison on the GSO dataset, in terms of PSNR, SSIM, LPIPS, Chamfer Distance (CD) $\times 10^{-3}$ and runtime efficiency. Notably, Time(g) and Time(r) denote the time of generating multi-view images and inputting these MVD images for generating rendered images, respectively. }
\label{compare1}
\end{table}
\subsection{Novel View Synthesis}
We evaluate the novel view synthesis quality of rendered per-view images compared with other methods.
The quantitative results are shown in Tab. \ref{compare1}.
Our method significantly outperforms all recent methods by a large margin across all view synthesis metrics.
The PSNR, SSIM, and LPIPS metrics for novel view synthesis on the GSO dataset are improved by 5.59dB, 0.063, and 0.064, respectively, compared to the second-best metrics.
It indicates that the rendered images of our method are more structurally similar to the ground truth.
We also provide qualitative results in Fig. \ref{fig:RGB_C}. 
It can be observed that the baseline methods usually yield inconsistent and irrational results.
For example, LGM \cite{tang2025lgm} and TriplaneGaussian \cite{zou2024triplane} may generate the flattened laptop (first row) and thick castle (second row).
It shows the difficulty in the direct learning of unstructured 3D Gaussians from 2D images.
Existing methods lack 3D spatial structures to effectively regulate the spatial distribution of 3D Gaussians, thereby limiting their ability to achieve a higher level of view consistency between rendered images and the input image.
Moreover, distorted geometric details and blurry textures are observed in recent methods, such as the worn-out and fuzzy bee toy (third row), shoes (fifth row) and teddy bear (sixth row).
These inconsistencies once again underscore the importance of effectively integrating RGB and normal images for the recovery of both geometric and semantic details.
Thanks to the 3D-native structure and efficient fusion of RGB and normal images, our method is capable of generating high-quality 3D objects exhibiting superior semantic and geometric consistency.
Please refer to the supplemental material for more results.

%

%
\begin{figure*}[!ht] \centering
	\includegraphics[width=1\linewidth]{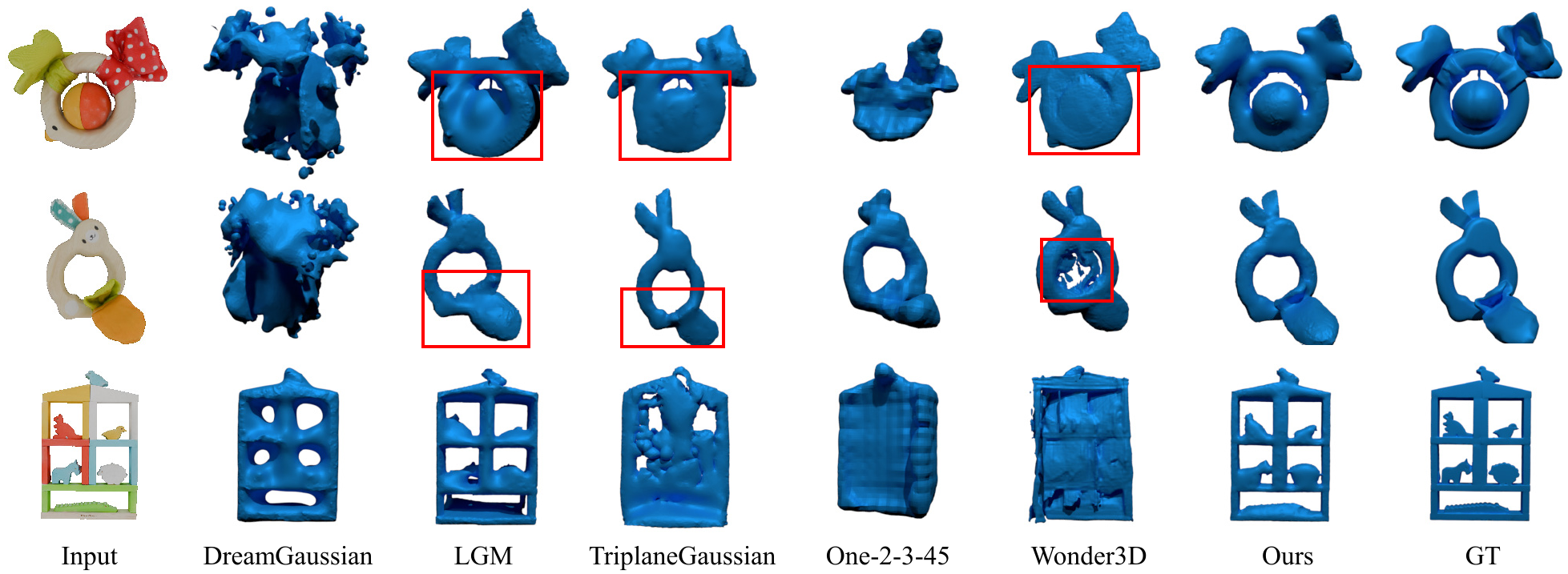}
	\caption{
Qualitative comparisons of single view reconstruction between GS-RGBN and other methods on the GSO dataset. 
    }
	\label{fig:GEO_C}
\end{figure*}
\begin{table}[!t]
\centering
\footnotesize
\begin{tabular}{c|ccc}
\toprule
Design           & PSNR$\uparrow$ & SSIM$\uparrow$ & LPIPS$\downarrow$ \\\midrule
Image-Gaussian&18.82&0.831&0.209\\
w/o LPIPS loss &21.83&0.859& 0.151     \\
w/o depth loss &21.62     &0.858     &0.154      \\
w/o regularization loss &22.51     &0.867     &0.142     \\
w/o normal input&20.15     &0.848     &0.172     \\
w/o CVF&19.27     & 0.843    &0.198    \\
w/o $CA_{s}$&21.08 &0.852&0.166\\
w/o $CA_{g}$&21.32&0.853&0.163\\
w/o $SA$&21.67&0.858&0.153\\
Full model&\textbf{23.02} &\textbf{0.873} &\textbf{0.135}  \\
\bottomrule
\end{tabular}
\caption{Ablation study on the different loss functions and normal fusion strategies on the GSO dataset.}
\label{ab}
\end{table}
\subsection{Single View Reconstruction}
We evaluate the single view reconstruction quality for different methods.
The quantitative and qualitative results are shown in Tab. \ref{compare1} and Fig. \ref{fig:GEO_C}.
It can be observed that the ambiguity of SDS leads to completely out-of-control 3D object generation like DreamGaussian \cite{tang2023dreamgaussian}. 
Both One-2-3-45 \cite{liu2024one} and Wonder3D \cite{long2024wonder3d} tend to generate meshes that are incomplete and distorted, particularly when it comes to preserving the mesh structures with holes.
LGM \cite{tang2025lgm} and TriplaneGaussian \cite{zou2024triplane} can generate shapes that exhibit rough alignment with the input image but fail to capture intricate details.
In contrast, our method uses the hybrid Voxel-Gaussian model to maintain geometry consistency between the generated shapes and ground truth and fully exploits geometric information from normal images to preserve finer geometric details.

\subsection{Runtime Efficiency}
We assess the runtime efficiency of GS-RGBN in comparison with other methods. 
For fair comparisons, we divide the total runtime into two components: the time for pre-trained MVD models to generate multi-view images (Time(g)), and the time for inputting these MVD images to produce rendered images using designed feed-forward models or the pre-shape optimization process (Time(r)).
Notably, the total runtime of DreamGaussian \cite{tang2023dreamgaussian} and TriplaneGaussian \cite{zou2024triplane} only contains Time(r).
As shown in Tab. \ref{compare1}, Gaussian-based feed-forward methods (TriplaneGaussian \cite{zou2024triplane}, LGM \cite{tang2025lgm} and GS-RGBN) exhibit significantly reduced rendering time compared to traditional approaches (Wonder3D \cite{long2024wonder3d} and One-2-3-45 \cite{liu2024one}) that utilize other 3D representations such as NeRF.
In particular, GS-RGBN demonstrates outstanding performance while still maintaining acceptable efficiency.
Given the superior performance achieved, it is deemed acceptable for our method to allocate additional time towards establishing a structured 3D voxel grid and aggregating more MVD RGB/normal images compared to TriplaneGaussian and LGM.

\begin{figure*}[!ht] \centering
	\includegraphics[width=1\linewidth]{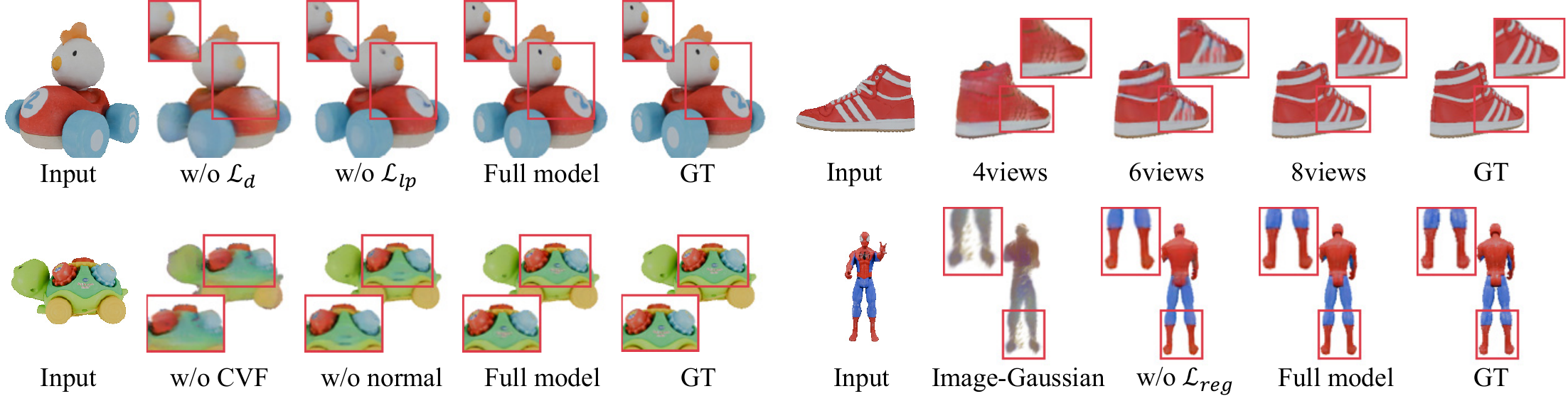}
	\caption{
Ablation study of different training models. Our full model achieves the best 3D object reconstruction with consistent details.
    }
	\label{fig:ab}
\end{figure*}
\subsection{Ablation study}
\textbf{Effect of Hybrid Voxel-Gaussian.} 
We conduct experiments to evaluate the effect of the hybrid Voxel-Gaussian model as shown in Table \ref{ab} (first row).
When we remove the process of constructing 3D feature volumes and directly feed the 2D RGB and normal feature maps into a modified 2D CVF module to encode the final Gaussians in an Image-Gaussian manner, similar to previous feed-forward methods \cite{tang2025lgm,xu2024grm,zhang2025gs,shen2024gamba}, there is a significant decline in model performance.
As shown in Fig. \ref{fig:ab}, removing the hybrid Voxel-Gaussian model makes it exceedingly challenging to control the movement and shape changes of 2D Gaussians. 
It implies that the hybrid Voxel-Gaussian representation is indispensable since it builds a structured 3D voxel grid, facilitating generalizable 3D learning of unstructured 2D Gaussians for recovering the geometric intricacies of 3D objects.

\textbf{Effect of Loss Functions.} 
The whole paradigm can be supervised by employing only the L1 loss between RGB and alpha images to ensure a fundamental training process, while we assess the effect of additional loss functions.
The model performance decreases when the LPIPS, depth, and regularization loss terms are successively removed, as demonstrated in Table \ref{ab}.
It means that all additional loss functions significantly enhance the overall quality of the reconstructed 3D object.
Especially, the depth and regularization loss functions, which cannot be achieved by 3D Gaussian-based methods due to varying depth values, can enhance texture quality (see Fig. \ref{fig:ab}).

\textbf{Effect of Normal Fusion.} 
We conduct experiments to evaluate the effect of normal fusion, as shown in Table \ref{ab}.
We first remove the input multi-view normal maps and observe that the performance significantly drops (ambiguous geometric intricacies in Fig. \ref{fig:ab}), demonstrating the indispensable role of normal images in providing geometric guidance and crucial clues for recovering intricate geometric details.
%
We propose to use CVF module to effectively aggregate RGB and normal volumetric features. 
As shown in Table \ref{ab}, we first remove the whole CVF module and directly concatenate RGB and normal volumetric features into MLPs. We observe a very significant performance drop, indicating that the CVF module offers an effective way of fusing RGB and normal information.
Besides, we replace cross-attention and self-attention blocks in CVF with simple average pooling layers, which is widely used for feature aggregation in previous multi-view stereo (MVS) approaches \cite{chen2021mvsnerf,long2022sparseneus}.
The observed decline in model performance suggests that attention-based mechanisms with varying attention weights offer greater benefits to cross-volume fusion.
Furthermore, we investigate the impact of voxel residual blocks (VRBs) in CVF, as presented in Table \ref{ab2}. The model performance demonstrates a decline when reducing the number of VRBs from 3 to 1 or substituting them with 3D CNNs, owing to the incorporation of encoded spatial features from VRBs.
\begin{table}[!t]
\centering
\footnotesize
\begin{tabular}{cc|ccc}
\toprule
VRBs&Views& PSNR$\uparrow$ & SSIM$\uparrow$ & LPIPS$\downarrow$ \\\midrule
Convs &8&22.04 & 0.860    & 0.150\\
1 &8     &19.49     &0.845    &0.195 \\
2 &8    &20.63     &0.854    &0.159 \\
3& 8    &21.76     &0.859   &0.153  \\  \midrule
4& 4    & 20.06    &0.848   &0.165  \\
4& 6    &22.70 &0.868& 0.141\\
4& 8&\textbf{23.02} &\textbf{0.873} &\textbf{0.135} \\
\bottomrule
\end{tabular}
\caption{Ablation study on different VRBs and source views on the GSO dataset. Convs means that we replace all four VRBs with standard 3D CNNs.}
\label{ab2}
\end{table}

\textbf{Effect of Different Views.} 
We train our paradigm with different input views from Wonder3D, as shown in Tab. \ref{ab2}. 
The model performance demonstrates a significant improvement as the number of input views increases, indicating that our model effectively integrates valuable information from more inconsistent MVD images to obtain better 3D reconstruction results (see Fig. \ref{fig:ab}).
Especially, given only four views (0, 90, 180, and 270 degrees azimuths), our model still surpasses existing methods in terms of the quality of generated 3D objects (refer to Table \ref{compare1} and Table \ref{ab2}).
%


\section{Conclusion and Limitations}
In this paper, we propose GS-RGBN, an RGBN-Volume Gaussian Reconstruction Model that enables fast and high-fidelity 3D object generation from single-view images.
Our method consists of two key components: the 3D-native hybrid Voxel-Gaussian model for structured 2D Gaussian learning, and the cross-volume fusion (CVF) module for effectively fusing RGB and normal information to ensure view-consistent geometric details.
Similar to existing single-view 3D reconstruction methods, GS-RGBN heavily relies on MVD models to generate multi-view RGB and normal images for 3D object generation.
The performance degradation occurs when the MVD models generate images with a higher level of view inconsistency.
The large-scale scene generation requires a large model pretrained for multi-view scene image generation. In the future, we also plan to pretrain an MVD model for multi-view image generation of large scenes, which could be used as a component of our method for large-scale scene generation. Besides, voxels cannot be directly used for representing large-scale scenes. We will explore using an octree of voxels~\cite{ren2024octree} to improve the memory efficiency for generating large-scale scenes.

\section*{Acknowledgments}
The authors would like to thank the reviewers for their insightful comments. This work is supported by NSF China (No. U23A20311, No. 62322209 and No. 62421003), the gift from Adobe Research, the XPLORER PRIZE, and the 100 Talents Program of Zhejiang University. The source code is available at
\textcolor{magenta}{ \href{https://gapszju.github.io/GS-RGBN}{https://gapszju.github.io/GS-RGBN}}.

%
%


%

%
{
    \small
    \bibliographystyle{ieeenat_fullname}
    \bibliography{main}
}


\end{document}